\documentclass[letterpaper, 10 pt, conference]{ieeeconf}  

\IEEEoverridecommandlockouts                              

\usepackage{header}
\usepackage{crs_acronyms}

\title{\LARGE \bf
Contextual Tuning of Model Predictive Control \\ for Autonomous Racing
}

\author{Lukas P. Fr\"ohlich$^{1, *}$, Christian K\"uttel$^{1}$, Elena Arcari$^{1}$, \\Lukas Hewing$^{1,2}$,  Melanie N. Zeilinger$^{1}$ and Andrea Carron$^{1}$
\thanks{$^{1}$The authors are \revised{members of} the Institute of Dynamic Systems and Control, ETH Zurich, Sonneggstrasse~3, 8092 Zurich, Switzerland}%
\thanks{$^{2}$L.\,H. is \revised{employed by} SENER Aerospace, Tres Cantos, Madrid, Spain}
\thanks{$^{*}$Corresponding author {\tt\small lukasfro@ethz.ch}}
\thanks{The authors would like to thank all students that contributed
to the racing platform, in particular, David Helm. Andrea
Carron’s research was supported by the Swiss National Centre
of Competence in Research NCCR Digital Fabrication and
the ETH Career Seed Grant 19-18-2.}%
}

\newcommand{\revised}[1]{\textcolor{black}{#1}}
\newcommand{\final}[1]{\textcolor{black}{#1}}

\begin{document}

\maketitle
\thispagestyle{empty}
\pagestyle{empty}


\begin{abstract}
      Learning-based model predictive control has been widely applied in autonomous racing to improve the closed-loop behaviour of vehicles in a data-driven manner.
      When environmental conditions change, e.g., due to rain, often only the predictive model is adapted, but the controller parameters are kept constant.
      However, this can lead to suboptimal behaviour.
      In this paper, we address the problem of data-efficient controller tuning, adapting both the model and objective simultaneously.
      The key novelty of the proposed approach is that we leverage a learned dynamics model to encode the environmental condition as a so-called context.
      This insight allows us to employ contextual Bayesian optimization to efficiently transfer knowledge across different environmental conditions.
      Consequently, we require fewer data to find the optimal controller configuration for each context.
      The proposed framework is extensively evaluated with more than 3'000 laps driven on an experimental platform with 1:28~scale RC race cars.
      The results show that our approach successfully optimizes the lap time across different contexts requiring fewer data compared to other approaches based on standard Bayesian optimization.
\end{abstract}

\section{Introduction}

In recent years, learning-based methods have become increasingly popular to address challenges in control for autonomous racing.
\Gls{mpc} is of particular interest for learning-based approaches as it can deal with state and input constraints and offers several ways to improve the closed-loop performance from data \cite{Hewing2020LearningMPC,Domahidi2011LearningMPC, Rosolia2020LearningRaceCar}.
A common method of using data to improve \gls{mpc} is by learning a predictive model of the open-loop plant from observed state transitions \cite{Bansal2017GoalDrivenBO, Kabzan2019learningMPC}.
However, as environmental conditions can change, it is imperative to adapt the learned model during operation instead of keeping it fixed \cite{McKinnon2019learnFastForgetSlwo}.

In addition to the predictive model, the effective behaviour of \gls{mpc} is governed by several parameters.
While for simple control schemes such as PID controllers, there exist heuristics to tune the parameters \cite{Ziegler1942optimumSettings}, this is generally not the case for \gls{mpc}. Therefore, practitioners often resort to hand-tuning, which is potentially inefficient and suboptimal.
In contrast, an alternative approach to controller tuning is to treat it as black-box optimization problem and use zeroth-order optimizers such as CMA-ES \cite{Hansen2001CMAES}, DIRECT \cite{Jones1993Direct} or \gls{bo} \cite{Garnett2022BayesOptBook} to find the optimal parameters.
Due to its strong sample efficiency and ability to cope with noise-corrupted objective functions, \gls{bo} has become especially popular \cite{Chatzilygeroudis2019PolicySearchHandfulTrials} and is therefore also employed in our approach.

In this paper, we argue that the optimal controller parameters need to be adapted when the environmental conditions change instead of keeping them fixed; akin to the need for adaptive model learning schemes.
Consider the following example for autonomous race cars:
During good weather conditions, the tires exhibit sufficient traction such that the controller can maneuver the car aggressively.
If it starts raining, the traction is reduced and the driving behaviour needs to be adapted accordingly, which renders the previous controller suboptimal.

There are several options to deal with the issue in the aforementioned example.
One option, for instance, is to assume that the optimal controller parameters change over time \cite{Bogunovic2016TimeVaryingBanditOptimization}.
This approach `forgets' old parameters over time and continues to explore the full parameter space over and over again as time progresses.
However, simply forgetting old data is inefficient if previously observed scenarios re-occur.
Alternatively, one can try to find a `robust' controller parameterization that is suited for several different environmental conditions \cite{Tesch2011SnakeGaitBOChangingEnvironments,Froehlich2020NoisyInputES}.
However, the drawback of this approach is that the robust controller configuration performs well on average, but is not necessarily optimal for any of the specific conditions.
In contrast, we aim to find the optimal controller for each context to obtain the optimal lap time in all conditions.

\textbf{Contributions }
The contributions of this paper are threefold:
Firstly, we propose to jointly learn a dynamics model and optimize the controller parameters in order to fully leverage the available data and efficiently adapt to new conditions.
In particular, we encode the environmental conditions via a learned dynamics model as context variables.
To this end, we derive a novel parametric model that is tailored to account for model inaccuracies imposed by lateral tire forces, which are critical for high-performance autonomous racing.
The context variables are then leveraged by contextual \gls{bo} to transfer knowledge between different contexts, which allows for more efficient controller tuning under unknown conditions.
Secondly, we demonstrate the effectiveness of our proposed framework with an extensive experimental evaluation on a custom hardware platform with 1:28 scale RC race cars totalling up to more than 3'000 driven laps.
To the best of our knowledge, this paper provides the first demonstration of contextual \gls{bo} for a robotic application on hardware.
Lastly, we provide an open-source implementation of (contextual) \gls{bo} for the \gls{ros} \cite{Quigley2009ros} framework at \linebreak\url{www.github.com/IntelligentControlSystems/bayesopt4ros}; the first of its kind in the official \gls{ros} package distribution list.

\section{Related Work}
\label{sec:related_work}
%
\subsection{Combining Model Learning and Controller Tuning}
In recent years, \gls{bo} has found widespread applications in control, robotics and reinforcement learning, owing its success to superior sample efficiency \cite{Chatzilygeroudis2019PolicySearchHandfulTrials}.
In this section, we focus on approaches that combine model learning with \gls{bo}.
The algorithm proposed in \cite{Wilson2014TrajectoryBO} computes a prior mean function for the objective's surrogate model by simulating trajectories with a learned model.
They account for a potential bias introduced due to model inaccuracies by means of an additional parameter governing the relative importance of the simulated prior mean function, which can be inferred through evidence maximization.
The work in \cite{Bansal2017GoalDrivenBO} builds on the idea of `identification for control' \cite{Gevers2005IdentifcationForControl} and uses \gls{bo} to learn dynamics models not from observed state transitions\revised{,} but in order to directly maximize the closed-loop performance of \gls{mpc} controllers.
Alternatively, the work in \cite{Froehlich2019domainSelectionBO} utilizes a learned dynamics model to find an optimal subspace for linear feedback policies, enabling \revised{application} to high-dimensional controller parameterizations.
Recently, two general frameworks have been proposed to jointly learn a dynamics model and tune the controller parameters in an end-to-end fashion \cite{OKelly2020TunerCar, Edwards2021AutoMPC}.
Both frameworks have only been validated in simulation and neither considers varying environmental conditions to adapt the controller parameters.

\subsection{Contextual Bayesian Optimization for Robotics}
The foundation for the work on contextual \gls{bo} was first explored in the multi-armed bandit setting \cite{Krause2011contextual}.
Therein, a variant of the \gls{ucb} acquisition function was proposed and a multi-task \gls{gp} \cite{Bonilla2008MultiTaskGP} was employed to model the objective as a function of both the decision variables and the context.
The idea of using context information to share knowledge between similar tasks or environmental conditions has found widespread applications in the field of robotics.
The authors in \cite{Metzen2015ContextualPSwithBO} applied contextual \gls{bo} to control a robotic arm with the aim of throwing a ball to a desired target location.
In particular, they optimize the controller parameters with \gls{bo} and consider the target's coordinates as context variables.
Another promising application of contextual \gls{bo} is locomotion for robots, wherein the context typically \revised{describes} the surface properties, such as inclination or terrain shape \cite{Yang2018ContextualBOLocomotionPrimitives,Seyde2019LocmotionContextBO}.
In \cite{Fiducioso2019SafeContextualBO}, a contextual variant of the SafeOpt algorithm \cite{Berkenkamp2016ControllerOptimization} is applied to tune a PID controller for temperature control where the outside temperature is considered \revised{to be a} context variable.
Related to contextual \gls{bo} is also the `active' or `offline' contextual setting \cite{Fabisch2014activeContextualPolicySearch}, wherein the context can be actively selected, e.g., in simulations or controlled environments.

\subsection{Model Predictive Control for Autonomous Racing}
\gls{mpc} is well understood \revised{theoretically} and allows for \revised{incorporation of} state and input constraints to safely race on a track, making it a prime candidate as an approach for controlling autonomous cars.
Two common approaches to the racing problem are time-optimal control \cite{Metz1989TimeOptimalRacing} and contouring control \cite{Lam2010MPCC}, i.e., where the progress is maximized along the track.
Both approaches have been implemented successfully on hardware platforms akin to ours \cite{Verschueren2014TimeOptimalMPC,Liniger2015RCCars}.
In order to account for modeling errors, several learning approaches have been proposed in the context of race cars.
The authors in \cite{Hewing2019cautiousMPC} use \gls{gp} regression to learn a residual dynamics model and employ an adaptive selection scheme of data points to achieve real-time computation.
A similar approach has been applied to a full-scale race car \cite{Kabzan2019learningMPC}.
Alternatively, the algorithm proposed in \cite{Rosolia2020LearningRaceCar} uses time-varying locally linear models around previous trajectories without relying on a nominal model.

\subsection{Other related work}
\revised{The following ideas are also} related to this paper, but do not directly fit into any of the three previous categories.
The authors in \cite{Wischnewski2019SafeOptRacecarTuning} employ a safe variant of \gls{bo} \cite{Berkenkamp2016ControllerOptimization} to increase the controller's performance without violating \revised{the prespecified} handling limits of the car.
The approach proposed in \cite{Jain2020racingLineBO} uses \gls{bo} to optimize the ideal racing line instead of relying on an optimal control formulation.
Lastly, the authors in \cite{McKinnon2020ContextAwareCostShaping} propose to learn the discrepancy between the predicted and observed \gls{mpc} cost in order to account for modeling errors of an autonomous vehicle in different contexts.

\section{Preliminaries}
\label{sec:preliminaries}

In this section, we describe the dynamics model \revised{for the vehicle} for which the tire forces are the prime source of modelling errors.
Subsequently, we review the \gls{mpc} scheme and discuss its relevant tuning parameters.
Lastly, we formalize the controller tuning problem as \revised{a} black-box optimization and how \gls{bo} can be employed to address it.

\subsection{Vehicle Model}
\label{sec:vehicle_model}

In this work, we model the race \revised{car} dynamics with the bicycle model~\cite{Rajamani2011Vehicle} (see \cref{fig:bicycle_model}).
The state of the model is given by $\state = \left[p_x, p_y, \psi, v_x, v_y, \omega \right]$, where $p_x$, $p_y$, $\psi$ denote the car's position and heading angle in the global coordinate frame, respectively.
The velocities and yaw rate in the vehicle's body frame are denoted by $v_x$, $v_y$, and $\omega$.
The car is controlled via the steering angle $\delta$ and the drive train command $\tau$, summarized as input $\action = [\delta, \tau]$.
The states evolve according to the \revised{nonlinear} discrete-time dynamics
\begin{align}\label{eq:discrete_model}
    \state_{\ti+1} = f(\state_\ti, \action_\ti) + g(\state_\ti, \action_\ti),
\end{align}
where $f$ denotes the nominal model and $g$ the residual model accounting for \revised{unmodelled} effects, which \revised{we will learn} from data.
The \revised{nominal model} is governed by the set of \revised{ODEs}
\begin{align}\label{eq:continous_model}
    \begin{bmatrix}
        \dot{v}_x \\ \dot{v}_y \\ \dot{\omega}
    \end{bmatrix} =
    \begin{bmatrix}
        \frac{1}{m} \left(F_x - F_{\mathrm{f}}\sin(\delta) + m v_y \omega \right)            \\
        \frac{1}{m} \left(F_{\mathrm{r}} + F_{\mathrm{f}}\cos(\delta) - m v_x \omega \right) \\
        \frac{1}{I_z} \left(F_{\mathrm{f}} l_{\mathrm{f}}\cos(\delta) - F_{\mathrm{r}} l_{\mathrm{r}} \right)
    \end{bmatrix},
\end{align}
with $m$ as the car's mass, $I_z$ being the moment of inertia, and $l_{\mathrm{f}/\mathrm{r}}$ define the distance between the center of gravity and the front and rear axles, respectively.
One of the most critical components in the bicycle model for the use in high-performance racing are the lateral forces acting on the tires.
To this end, we employ the simplified Pacejka model~\cite{Pacejka2002Tire}
\begin{align}\label{eq:pacejka_model}
    F_{\mathrm{f}/\mathrm{r}} = D_{\mathrm{f}/\mathrm{r}} \sin(C_{\mathrm{f}/\mathrm{r}} \arctan(B_{\mathrm{f}/\mathrm{r}} \alpha_{\mathrm{f}/\mathrm{r}}))
\end{align}
for the front (f) and rear (r), respectively.
The parameters $B, C$, and $D$ are found via system identification and depend on the car itself as well as on the friction coefficient between the tires and road surface.
The slip angles $\alpha_{\mathrm{f}/\mathrm{r}}$ are defined as
\begin{align*}
    \alpha_\mathrm{f} = \arctan \left( \frac{v_y + l_{\mathrm{f}} \omega}{v_x} \right) - \delta, \hspace{0.5em} \alpha_\mathrm{r} = \arctan \left( \frac{v_y - l_{\mathrm{r}} \omega}{v_x} \right).
\end{align*}
For driving maneuvers at the edge of the car's handling limits, it is paramount to estimate the lateral tire forces as accurately as possible.
To this end, we derive a parametric model tailored to account for inaccuracies \revised{in} the tire model \cref{eq:pacejka_model} in \cref{sec:model_learning}.

\begin{figure}[t]
	\centering
	\begin{tikzpicture}
		\tikzset {
			wheel/.style n args={3}{
					rectangle,
					fill,
					black,
					minimum width={#1},
					minimum height={#2},
					rotate={#3},
					rounded corners,
				}};
		\def\height{4cm};
		\def\radius{\height/5};
		\def\width{\height/15};
		\def\cor{\height/1.2};
		\def\steer{30};
		\def\bet{atan(\height/2/(\cor))};
		\def\arcradius{0.8};
		\def\yaw{20};
		\def\lengthsep{0.8};
		\def\dynang{(\bet)};

		\begin{scope}[rotate=\yaw]
			\node[wheel={\radius}{\width}{\yaw}] (wr) at (-\height/2,0){};
			\node[wheel={\radius}{\width}{\steer+\yaw}] (wf) at (\height/2,0){};
			\node[draw,inner sep=\height/50, circle,radius = \height/50,fill, label=below:{$(x, y)$}] (cog) at (0,0){};

			\draw[dashed] (wf) -- ++({\steer}:1);
			\draw[dashed] (wf) -- ++(1,0);
			\draw[->] (wf.center) ++(0:\arcradius) arc (0:\steer:\arcradius) node[midway,shift={(0.15,0.15)}]{$\delta$};

			\def\vell{1.5};
			\draw (wr) -- (wf);
			\draw[->,blue,thick] (cog) -- ++({\bet}:\vell) node()[shift={(-0.2,0)}]{$v$};
			\draw[->] (cog) ++(0:\arcradius) arc (0:{\bet}:\arcradius) node[anchor = west,shift={(0.1,0)}]{$\beta$};

			\draw[->,blue,thick] (cog) -- ++(0,{sin(31)*\vell} ) node()[shift={(-0.1,-0.4)}]{$v_y$};
			\draw[->,blue,thick] (cog) -- ++({cos(31)*\vell},0 ) node()[shift={(0,0.2)}]{$v_x$};

			\draw[->,blue,thick] (wr.center) -- ++({\dynang}:1);
			\draw[->] (wr.center) ++(0:\arcradius) arc (0:\dynang:\arcradius) node()[shift={(0.4,0.0)}]{$\alpha_\mathrm{r}$};

			\draw[->,blue,thick] (wf.center) -- ++({\dynang+\steer}:1);
			\draw[->] (wf.center) ++(\steer:\arcradius) arc (\steer:{\dynang+\steer}:\arcradius) node()[shift={(0.35,0.1)}]{$\alpha_\mathrm{f}$};

			\draw[->,red,thick] (wr) -- node()[anchor = east]{$F_\mathrm{r}$} ++(0,\arcradius);
			\draw[<-,red,thick] (wr) -- node()[anchor = north]{$F_x$} ++(-1.5*\arcradius,0);
			\draw[->,red,thick] (wf) -- node()[shift={(0.1,0.2)}]{$F_\mathrm{f}$} ++({\steer+90}:\arcradius);
            
			\draw[|-|] ($(wr)+(0,-\lengthsep)$) -- node()[anchor = north]{$l_\mathrm{r}$} ($(cog)+(0,-\lengthsep)$);
			\draw[|-|] ($(cog)+(0,-\lengthsep)$) -- node()[anchor = north]{$l_\mathrm{f}$} ($(wf)+(0,-\lengthsep)$);

			\def\psiangle{60};
			\draw[->] (cog.center) ++(\psiangle:1) arc (\psiangle:180-\psiangle:1) node[midway,shift={(0,0.3)}]{$\omega$};
		\end{scope}

		\draw[dashed] (cog.center) -- ($(cog.center)+(1,0)$);
		\draw[->] (cog) ++(0:\arcradius) arc (0:\yaw:\arcradius) node[shift={(0.3,-0.1)}]{$\psi$};

		\def\loff{3cm};
		\def\doff{2cm};

	\end{tikzpicture}
	\caption{
		The dynamic bicycle model with corresponding velocities (blue) and forces (red).
		The inputs are given by the longitudinal drive train command $\tau$ acting on $F_x$ and the steering angle $\delta$.
		The tire forces $F_{\mathrm{f}/\mathrm{r}}$ are computed with the simplified Pacejka model based on the slip angles $\alpha_{\mathrm{f}/\mathrm{r}}$.
	}
	\label{fig:bicycle_model}
\end{figure}

\subsection{Model Predictive Contouring Control}\label{subsec:model_predictive_contouring_control}
In this section, we briefly review the \gls{mpcc} formulation for race cars introduced in \cite{Liniger2015RCCars}.
The goal in \gls{mpcc} is to maximize the progress along the track while satisfying the constraints imposed by the boundaries of the track as well as the car's dynamics.
The control problem is formalized by the \revised{nonlinear} program
\begin{align}
    \begin{split}
        \label{eq:mpcc}
        \action^* & = \argmin_{[\paction_0, \dots, \paction_{H}]} \sum_{i=0}^{H} \bm{\varepsilon}_i^\top \Q \bm{\varepsilon}_i - Q_{\mathrm{adv}} \gamma_{i} \\
        \text{s.t. } & \pstate_0 = \state_\ti, \; \pstate_{i+1} = f(\pstate_i, \paction_i) + g(\pstate_i, \paction_i)  \\
        & \pstate_i \in \mathcal{X}, \; \paction_i \in \mathcal{U},\; \gamma_i > 0,
    \end{split}
\end{align}
where $\bm{\varepsilon}$ denotes the lag and contouring error, respectively, and~$\gamma$ is the advancing parameter.
We solve an approximation to the above optimization problem using the state-of-the-art solver ForcesPRO~\cite{Domahidi2014ForcesPro,Zanelli2017ForcesNlp} in a receding horizon fashion\revised{. The} action $\action_\ti$ for a given state $\state_\ti$ at time step $\ti$ is \revised{then} given by the first element of the optimal action sequence~$\action^*$.
The parameters $\Q = \operatorname{diag}[Q_{\mathrm{lag}}, Q_{\mathrm{cont}}]$ and $Q_{\mathrm{adv}}$ weigh the relative contribution of the cost terms and determine the effective driving behaviour of the car.
\revised{A benefit of the \gls{mpcc} approach over tracking control is that is allows simultaneous path planning and tracking with a learned dynamics model in real-time. Whereas in tracking control, the reference trajectory is computed offline and not adapted for the learned dynamics.}

\emph{Remark:}
Note that the objective in \cref{eq:mpcc} does not directly encode the closed-loop performance of interest, i.e., the lap time.
\revised{In particular, we do not know a priori} \revised{which} values \revised{for} the cost parameters lead to the best \revised{lap time} and thus require careful tuning.
While a time-optimal formulation would not necessarily require a similar tuning effort, the optimization problem itself is significantly more complex and real-time feasibility becomes an issue~\revised{\cite{liniger2015cars}}.

\subsection{Controller Tuning via Bayesian Optimization}
\label{sec:controller_tuning}
We assume that a controller is parameterized by a vector $\bovar \in \bodomain \subseteq \R^\nbovar$ \revised{that} influences the controller's performance with respect to some metric \mbox{$J: \bodomain \rightarrow \R$}, for instance, \final{how the cost parameters $\Q$ in \cref{eq:mpcc} affect the lap time.}
The controller tuning problem \revised{is then formalized as}
\begin{align}\label{eq:controller_tuning_problem}
    \bovar^*  = \argmin_{\bovar \in \bodomain} J(\bovar).
\end{align}
The optimization problem \cref{eq:controller_tuning_problem} has three caveats, which make it difficult to solve:
1) No analytical form of $J$ exists, but we can only evaluate it pointwise.
2) Each observed function value is typically corrupted by noise.
3) Each function evaluation corresponds to a full experiment that has to be run on hardware, \revised{which is why sample-efficiency is paramount.}
These \revised{caveats} render \revised{most} numerical optimization methods impractical and we resort to \acrfull{bo}, a sample-efficient optimization algorithm designed to \revised{address} stochastic black-box functions as described above \cite{Garnett2022BayesOptBook}.

\gls{bo} works in an iterative manner and has two key ingredients: a probabilistic surrogate model approximating the objective function and \revised{an} acquisition function $\acqfunc(\bovar)$ to choose new evaluation points.
A common choice for the surrogate model are \glspl{gp} \cite{Rasmussen2006Book}, which allow for closed-form inference of the posterior mean and variance based on \revised{noisy function values} $\databo_n = \{ \bovar_i, y_i = J(\bovar_i) + \epsilon_i\}_{i=1}^{n}$ \revised{ with,
    \begin{align}
        \begin{split}
            \label{eq:gp_predictive_distribution}
            \mu_n(\bovar) = \bm{k}_n^\top \bm{K}^{-1} \bm{y}, \; \sigma_n^2(\bovar) = k(\bovar, \bovar) - \bm{k}_n^\top \bm{K}^{-1} \bm{k}_n,
        \end{split}
    \end{align}
    with kernel function $k(\cdot\,|\,\cdot)$, $[\bm{k}_n]_i = k(\bovar, \bovar_i)$, $[\bm{K}]_{ij} = k(\bovar_i, \bovar_j) + \delta_{ij}\sigma_\epsilon^2$, $[\bm{y}]_i = J(\bovar_i) + \epsilon_i$, and $\delta_{ij}$ denotes the Kronecker delta.
    Given the predictive distribution in \cref{eq:gp_predictive_distribution}, }
the next evaluation point $\bovar_{n+1}$ is then chosen by maximizing the acquisition function\revised{, which trades off between exploration and exploitation.}
In this paper, we \revised{employ} the \gls{ucb} acquisition function given by $\acqfunc_{\textsc{ucb}}(\bovar | \databo_n) = \mu_n(\bovar) + \beta \sigma_n(\bovar)$ with \revised{$\beta = 2$} as exploration parameter.
\revised{The \gls{bo} algorithm then alternates between choosing new parameters and updating the surrogate model with new function evaluations.}

\section{Proposed Learning Architecture}\label{sec:learning_architecture}

In the following, we describe the two components that form the core contribution of this paper: learning the dynamics model of the cars using custom feature functions and the contextual controller tuning based on \gls{bo}.

\subsection{Learning the Dynamics Model}\label{sec:model_learning}

The goal of learning the residual dynamics \revised{$g$} is to account for any effects that are not captured by the nominal model~$f$ to improve the predictive performance of the full model~\cref{eq:discrete_model}.
Especially for controllers that rely on propagation of the model, such as \gls{mpc}, the predictive accuracy strongly impacts the control performance.
However, there is a trade-off to be struck between the learned model's accuracy and complexity as the full control loop has to run in real time.
To this end, we use a linear model with \revised{nonlinear} feature functions $\feature(\state, \action) = [\phi_1(\state, \action), \dots, \phi_{n_\phi}(\state, \action)]^\top : \R^{n_x} \times \R^{n_u} \rightarrow \R^{n_\phi}$ to regress the residual prediction errors \revised{of the $i$-th state} via
\begin{align}
    \label{eq:residual_model}
    g_i(\state, \action) = \w_i^\top \feature(\state, \action): \R^{n_x} \times \R^{n_u} \rightarrow \R,
\end{align}
\revised{with $\w_i \in \R^{n_\phi}$ as regression coefficients}.
\revised{The main benefit of the linear model is its computational efficiency.}
While the expressiveness of a linear model is \revised{determined} by its feature functions, \revised{it does not require a complex management of data points to retain the real-time capability of the controller, as opposed to a \revised{nonparametric} model such as a \gls{gp} \cite{Kabzan2019learningMPC}.}

For the considered application of autonomous racing, we only learn the residual dynamics of a subset of the states.
Notably, the change in position and orientation can readily be computed via integration of the respective time derivatives.
Further, we do not account for modeling errors of the longitudinal velocity $v_x$ as we found empirically that the nominal model was already sufficient for good predictive performance.
Consequently, the learned model only accounts for the lateral velocity $v_y$ and the yaw rate $\omega$.

To infer the regression coefficients $\w_i$, we opt for \gls{blr} \cite[Ch.~3.3]{Bishop2006Book}.
Utilizing a Bayesian treatment instead of ordinary linear regression has two advantages:
\revised{Firstly,} it leads to more robust estimates due to regularization via a prior belief over the coefficients.
\revised{Secondly, a probabilistic approach} includes uncertainty quantification \revised{and allows the controller to account for the uncertainty in the model prediction akin to \cite{Hewing2019cautiousMPC}}.

There is a wide variety of general-purpose feature functions~$\feature$, such as polynomials, radial basis functions, or trigonometric functions.
\revised{In this paper,}
we construct feature functions that are specific to the dynamic bicycle model.
In particular, we assume that the discrepancy in the predictive model is due to misidentified physical parameters~$\parameter$, such as mass, inertia and the parameters of the Pacejka tire model, i.e., \mbox{$\parameter_{\mathrm{true}} = \parameter_{\mathrm{nom.}} + \Delta \parameter$}.
\revised{For small errors in the nominal parameters, we can approximate the nominal model by
    \begin{align}
        \label{eq:pacejka_taylor_approximation}
        f_i(\state, \action; \parameter_\mathrm{true}) \approx f_i(\state, \action; \parameter_\mathrm{nom.}) + \jac_{\parameter} f_i(\state, \action; \parameter) \Delta \parameter,
    \end{align}
    with $\jac_{\parameter} f_i$ denoting the Jacobian of nominal model's $i$-th state with respect to $\parameter$.
}
Empirically, we found that the inertia~$I_z$ and the parameters $D_{\mathrm{f}/\mathrm{r}}$ in the Pacejka tire model~\cref{eq:pacejka_model} have the largest influence on the model's prediction error (we use the inverse inertia to retain linearity in the parameter).
\revised{With $\parameter = [I_z^{-1}, D_{\mathrm{f}}, D_{\mathrm{r}}]^\top$, the Jacobian for the lateral and angular velocities then becomes
\begin{align*}
    \jac_{\parameter} \begin{bmatrix}
        f_{vy} \\ f_{\omega}
    \end{bmatrix} =
    \begin{bmatrix}
        0                                      & \frac{1}{m D_\mathrm{f}} F_\mathrm{f} \cos(\delta)               & \frac{1}{m D_\mathrm{r}} F_\mathrm{r}              \\
        l_\mathrm{f} F_\mathrm{f} \cos(\delta) & \frac{l_\mathrm{f}}{I_z D_\mathrm{f}} F_\mathrm{f}  \cos(\delta) & \frac{l_\mathrm{r}}{I_z D_\mathrm{r}} F_\mathrm{r}
    \end{bmatrix}.
\end{align*}
Note that via the lateral tire forces $F_\mathrm{f}$ and $F_\mathrm{r}$, all elements in the Jacobian matrix are directly proportional to either of the following two terms: $\sin(C_\mathrm{f} \arctan (B_\mathrm{f} \alpha_\mathrm{f})) \cos(\delta)$ and $\sin(C_\mathrm{r} \arctan (B_\mathrm{r} \alpha_\mathrm{r}))$.}
\revised{
    Consequently, we can equivalently write the first-order term in \cref{eq:pacejka_taylor_approximation} as
    \begin{align*}
        \jac_{\parameter} f_i(\state, \action; \parameter) \Delta \parameter =
        \w_i^\top
        \underbrace{
            \begin{bmatrix}
                \sin(C_{\mathrm{f}} \arctan(B_{\mathrm{f}} \alpha_{\mathrm{f}})) \cos(\delta) \\
                \sin(C_{\mathrm{r}} \arctan(B_{\mathrm{r}} \alpha_{\mathrm{r}}))
            \end{bmatrix}}_{\feature_{\mathrm{Taylor}}(\state, \action)},
    \end{align*}
    which, when compared to \cref{eq:residual_model}, directly leads to what we call Taylor features $\feature_{\mathrm{Taylor}}(\state, \action)$.}

\revised{As the feature functions themselves are fixed, the shape of~$g$ is determined by the regression coefficients.
For instance, if the nominal model overestimates the tire friction, the residual model accounts for the mismatch, which leads to nonzero entries in $\w_i$.
In contrast, if the nominal model underestimates the tire friction, we would expect the regression coefficients’s values to be different.
Consequently, we can encode the environmental condition via the regression coefficients and summarize these as context $\context = [\w_{vy}^\top, \w_{\omega}^\top]^\top \in \R^4$.}


\subsection{Contextual Controller Tuning}
\label{sec:contextual_controller_tuning}
In \cref{sec:controller_tuning}, we have discussed how standard \gls{bo} can be employed to address the controller tuning problem by posing it as a black-box optimization.
This section \revised{discusses contextual \gls{bo} \cite{Krause2011contextual}, which enables the transfer of information between different environmental conditions.}

In order to account for the environmental condition encoded in the context variable $\context$, we extend the parameter space of the objective function $J$ such that it additionally depends on the context, $J: \bodomain \times \R^{\ncontext} \rightarrow \R$.
Consequently, the \gls{gp} \revised{surrogate model needs to include} the context variable as well.
We follow \cite{Krause2011contextual} and split the kernel for the joint parameter space into a product of two separate kernels such that
\begin{align*}
    k([\bovar, \context], [\bovar', \context']) = k_{\bovar}(\bovar, \bovar') k_{\context}(\context, \context').
\end{align*}
In this way, we can leverage \revised{the kernel approach} by encoding prior knowledge in the choice of the respective kernels.
With \revised{the} extended \gls{gp} model, we can generate a \revised{surrogate} model \revised{that accounts for} different contexts \revised{using all previous observations} $\databo_n = \{ \bovar_i, \context_i, J(\bovar_i; \context_i) + \epsilon_i \}_{i=1}^n$.

\begin{algorithm}[t]
    \begin{algorithmic}[1]
        \State $\databo_0 \gets$ Initialize BO \revised{with a space-filling design}
        \State \textcolor{Green}{$\datatel_0 \gets$ Collect initial telemetry data}
        \For{$n = 1 \leq N$}
        \State \textcolor{Green}{$\context_n \gets$ Context from BLR using $\datatel_{n-1}$}
        \State $\bovar_n \gets$ Optimize acquisition function $\acqfunc(\bovar | \databo_{n-1}, \textcolor{Green}{\context_{n}})$
        \State \textcolor{Green}{Update residual dynamics model \cref{eq:residual_model} with $\context_n$}
        \State Update \gls{mpcc} \cref{eq:mpcc} with parameters $\bovar_n$
        \State Drive extra \revised{lap} to mitigate transient effects
        \State $J_n, \textcolor{Green}{\datatel_{n}} \gets$ Observe cost \textcolor{Green}{and telemetry data}
        \State $\databo_{n} \gets \databo_{n-1} \cup (\bovar_n, \textcolor{Green}{\context_n}, J_n)$ Extend BO data
        \EndFor
    \end{algorithmic}
    \caption{Contextual \gls{bo} for controller tuning. Steps \revised{additional to} standard \gls{bo} are highlighted in \textcolor{Green}{green}.}
    \label{alg:contextual_bo}
\end{algorithm}

\cref{alg:contextual_bo} summarizes contextual \gls{bo} \revised{for controller tuning. We highlight the additional instructions for the contextual setting} in green.
In Line~4, the context $\context_n$ is \revised{inferred with \gls{blr}} from the \revised{current} telemetry data \revised{$\datatel_{n-1}$} as the regression coefficients of the residual model.
Since the context is governed by the environment itself we do not have influence over the respective values.
Accordingly, \revised{we only optimize the acquisition function} with respect to $\bovar$ in Line~5, while $\context_n$ is kept constant.
That way, \gls{bo} continues to explore the parameter space when new contexts are observed due to the increased predictive uncertainty of the \gls{gp}.
In contrast, for contexts that are similar to the ones encountered in previous \revised{iterations}, \gls{bo} readily transfers the experience and thus accelerates optimization.
\revised{In Lines~6--8, we update the dynamics model and the controller with $\context_n$ and $\bovar_n$, respectively, and drive an additional lap to mitigate any transient effects due to the new parameters.
    Both datasets, $\databo_n$ and $\datatel_n$, are then updated in Lines~9--10.}
Continuously updating the residual dynamics model and utilizing it as context variable allows us to account for either slowly changing environmental conditions, e.g., an emptying fuel tank or tires wearing \revised{out}, or deliberate changes \revised{in} the car's condition, such as new tires.
\revised{When a sudden environmental change occurs during a lap, e.g., due to rain, the car needs to drive an additional lap with the same controller parameters because the inferred context would otherwise be ambiguous due to misidentified model parameters.}
\revised{For a specific (increasing) sequence of \gls{ucb}'s exploration parameter $\beta_n$ and under certain regularity assumptions on $J$, \cref{alg:contextual_bo} exhibits sublinear contextual regret \cite{Krause2011contextual}.}

\section{Experiments}
\label{sec:experiments}
%

\usetikzlibrary{arrows}
\usetikzlibrary{arrows.meta}

\tikzstyle{arrowslow} = [->, dashed, red, thick, -{Triangle[round, length=2mm, width=1.5mm]}]
\tikzstyle{arrowfast} = [->, blue, thick, -{Triangle[round, length=2mm, width=1.5mm]}]

\tikzstyle{component} = [rectangle, minimum width=2.0cm, minimum height=1cm,text centered, fill=black!20, text width=1.8cm]

\begin{figure}[t]
    \centering
    \footnotesize
    \begin{tikzpicture}

        \node (model) [component]  at (1.05, 0.5) {Model learning \\ (\cref{sec:model_learning})};
        \node (cbo) [component]  at (1.05, 2.5) {Contextual BO \\ (\cref{sec:contextual_controller_tuning})};
        \node (ekf) [component]  at (4.05, 0.5) {Extended \\ Kalman Filter};
        \node (mpcc) [component]  at (4.05, 2.5) {MPCC \\ (\cref{subsec:model_predictive_contouring_control})};

        \node (car) []  at (7.8, 1.5) {\includegraphics[width=1.6cm]{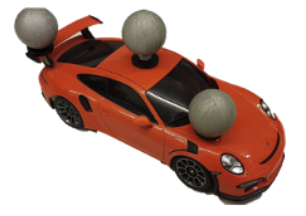}};
        \node (camera) []  at (6.5, 0.5) {\includegraphics[width=1.0cm]{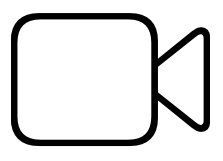}};
        \node (remote) []  at (6.5, 2.5) {\includegraphics[width=0.7cm]{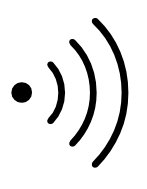}};

        \draw[arrowslow] (model) -- (cbo) node[midway, left] {$\context_n$};
        \draw[arrowslow] (cbo) -- (mpcc) node[midway, above] {$\bovar_n$};
        \draw[arrowfast] (ekf) -- (model) node[midway, above] {$\state_\ti$};
        \draw[arrowfast] (camera) -- (ekf) node[midway, above] {$\observation_\ti$};
        \draw[arrowfast] (mpcc) -- (remote) node[midway, above] {$\action_\ti$};

        \draw[arrowfast] ([xshift=-0.1cm]mpcc.south) -- ([xshift=-0.1cm]ekf.north) node[midway, left] {$\action_\ti$};
        \draw[arrowfast] ([xshift=0.1cm]ekf.north) -- ([xshift=0.1cm]mpcc.south) node[midway, right] {$\state_\ti$};

        \draw[arrowfast] ([xshift=0.14cm]mpcc.south west) -- ([xshift=0.14cm]model.north east) node[midway, below,xshift=0.2cm] {$\action_\ti$};
        \draw[arrowslow] ([xshift=-0.14cm]model.north east) -- ([xshift=-0.14cm]mpcc.south west) node[midway, above,xshift=-0.2cm] {$\context_n$};
    \end{tikzpicture}

    \caption{Schematic overview of the control and learning architecture with their respective communication channels. The subsystems are explained in \cref{sec:preliminaries,sec:learning_architecture}. \revised{The blue arrows denote communication in real-time at 35~Hz for the control loop. The dashed, red arrows denote the communication between the learning components, which occurs once every two laps.}
    }
    \label{fig:learning_architecture}
\end{figure}
We begin the experimental section by giving \revised{an overview} about the hardware platform and how the different components of the \revised{framework} interact with each other.
Subsequently, we evaluate the \revised{Taylor features in terms of predictive accuracy and the amount of required data.
    Lastly, we compare our framework with three approaches based on standard \gls{bo}.}

\revised{Throughout the experiments, we evaluate our framework on four different contexts to emulate different driving conditions.
    For contexts 1--2 and 3--4, we use two different cars and for contexts 2 and 4, we attach a mass of 40 grams (corresponding to around 30\% of the cars' total weight) to the cars' chassis.
    The additional mass influences the inertia, traction and steering behaviour of the cars.
}

\subsection{The Racing Platform}
\label{sec:racing_platform}
A schematic overview of the components and their respective communication channels is depicted in \cref{fig:learning_architecture}.
At the heart of the control loop is the \gls{mpcc} that computes the new input $\action_\ti$ for a given state $\state_\ti$ at 35 Hz; the respective input is sent via remote control to the race cars.
The cars are equipped with reflective markers enabling high-accuracy position and orientation measurements $\observation_{\ti}$ via a motion capture system, from which the full state $\state_{\ti}$ is estimated via an \gls{ekf}.
\revised{The component for model learning stores the telemetry data $\datatel$ for the last ten seconds in a first-in first-out buffer.
    Every two laps, the inferred context $\context_n$ is sent to the contextual \gls{bo}, which in turn computes the new controller parameters $\bovar_n$ for the \gls{mpcc}.}

\revised{For model learning, we require accurate estimates of the prediction error $\error_\ti = \state_{\ti+1} - f(\state_\ti, \action_\ti)$.
    However, the accuracy of the \gls{ekf}'s estimate for the velocity states is not sufficient for computing the prediction error because the estimator itself depends on the model.
    Since we only update the regression coefficients every two laps in \cref{alg:contextual_bo}, we can compute the prediction error in a `semi-offline' manner.
    To this end, we use the raw position and orientation measurements from the telemetry data $\datatel$} and compute the respective time derivatives numerically.
Since this \revised{computation} is done in hindsight, we can \revised{employ} non-causal filtering techniques to get a reliable ground truth for the velocity states.
While this approach leads to accurate state estimates, \revised{it adds} a time delay of a few seconds.
\revised{For our application, this does} not pose any problems in practice.

\subsection{Dynamics Learning}\label{sec:experiments_dynamics_learning}
\begin{figure}[t]
    \centering
    \includegraphics[scale=0.95]{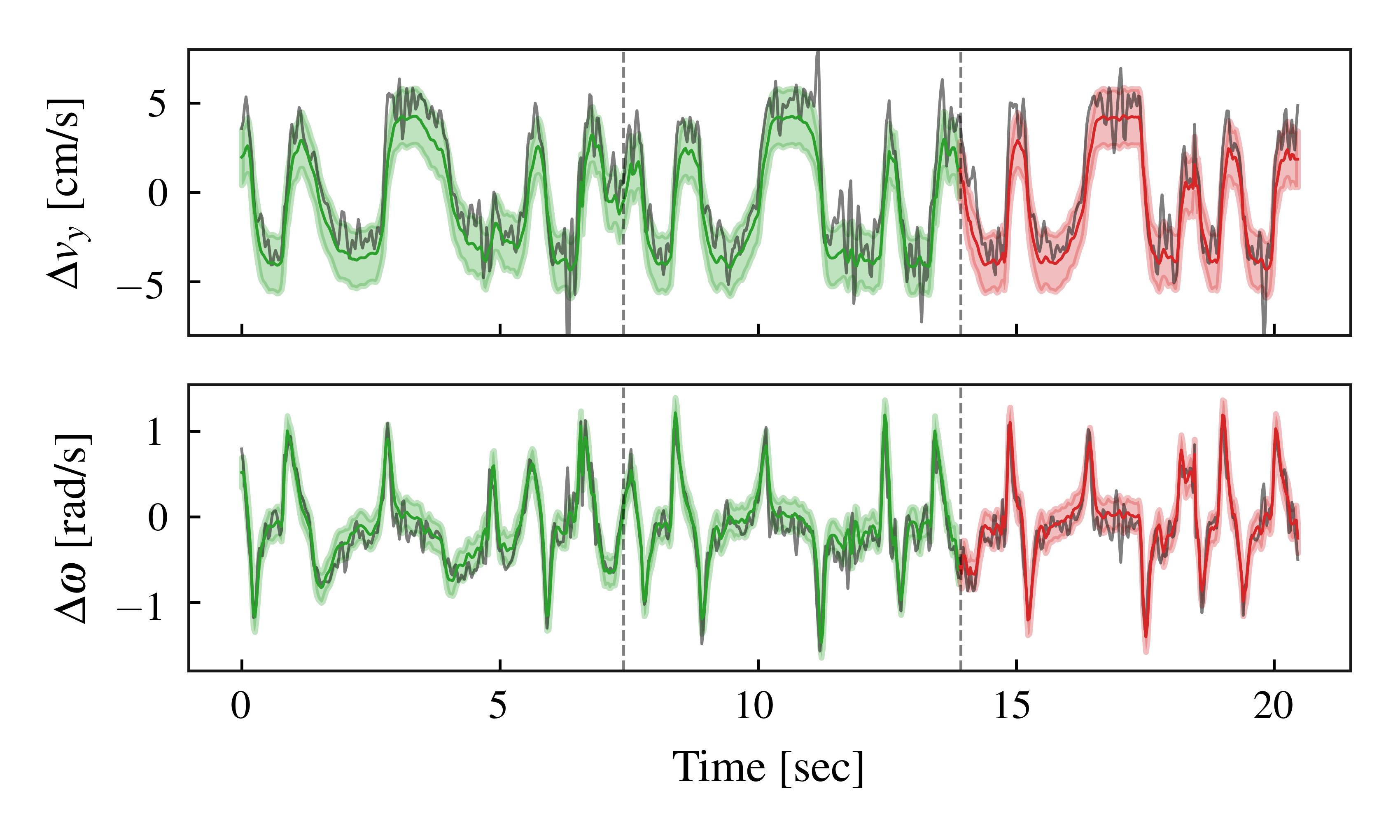}
    \vspace{-0.7cm}
    \caption{Error prediction (mean $\pm$ one standard deviation) from the learned residual model with $\bm{\phi}_{\mathrm{Taylor}}$ \cref{eq:pacejka_taylor_approximation} on training (green) and testing (red) data for lateral velocity (top) and yaw rate (bottom), respectively. The vertical dashed lines indicate the start of a new lap.}
    \label{fig:learning_error_signal}
\end{figure}

\begin{table}[t]
    \caption{State prediction errors for different learning models}
    \label{tab:prediction_error}
    \begin{tabularx}{\columnwidth}{cc|YYY}
        \toprule
        \multicolumn{2}{c|}{Prediction error} & No model learning & Gaussian Process & BLR with $\bm{\phi}_{\text{Taylor}}$                            \\
        \midrule
        \multirow{2}{*}{$v_y$ {[}cm/s{]}}     & train             & $3.81 \pm 0.21$  & $\mathbf{0.82 \pm 0.04}$             & $1.66 \pm 0.09$          \\
                                              & test              & $3.85 \pm 0.26$  & $\mathbf{1.07 \pm 0.14}$             & $1.77 \pm 0.05$          \\
        \midrule
        \multirow{2}{*}{$\omega$ {[}rad/s{]}} & train             & $0.46 \pm 0.03$  & $0.21 \pm 0.02$                      & $\mathbf{0.18 \pm 0.02}$ \\
                                              & test              & $0.44 \pm 0.02$  & $0.22 \pm 0.01$                      & $\mathbf{0.18 \pm 0.01}$ \\
        \bottomrule
    \end{tabularx}
\end{table}

\begin{figure*}
    \centering
    \includegraphics[width=0.95\linewidth]{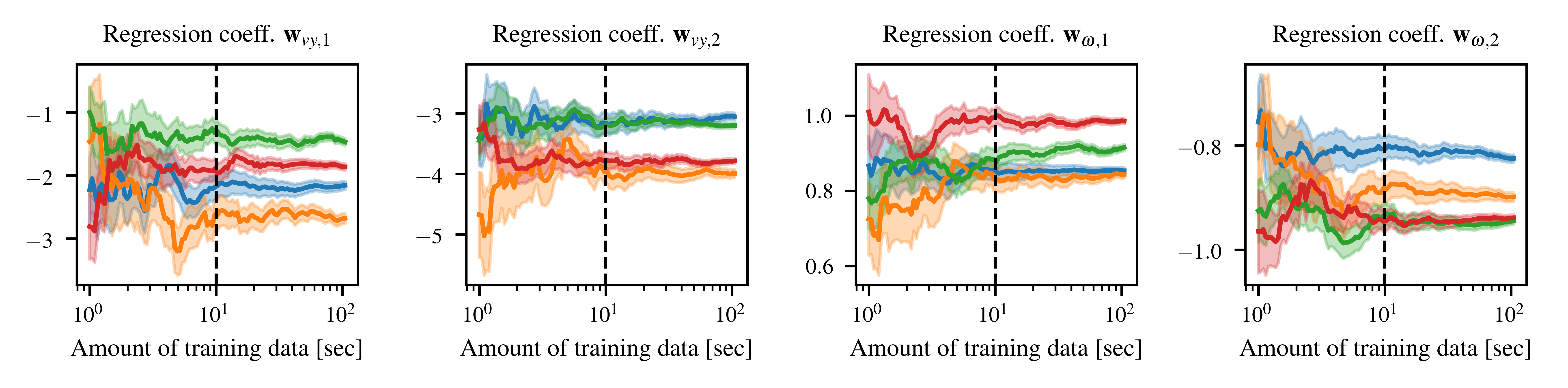}
    \vspace{-0.6cm}
    \caption{\revised{Inferred regression coefficients as a function of the amount of training data (in seconds) for the different contexts (\protect\bostandardlegendentry: context 1, \protect\bopriormeanlegendentry:~context~2, \protect\bocontextuallegendentry: context 3, \protect\bocumulativelegendentry: context 4). The solid lines depict the posterior mean estimate and the shaded area corresponds to mean $\pm$ one standard deviation.}}
    \label{fig:number_of_data}
\end{figure*}
\begin{figure*}
    \centering
    \includegraphics[scale=0.95]{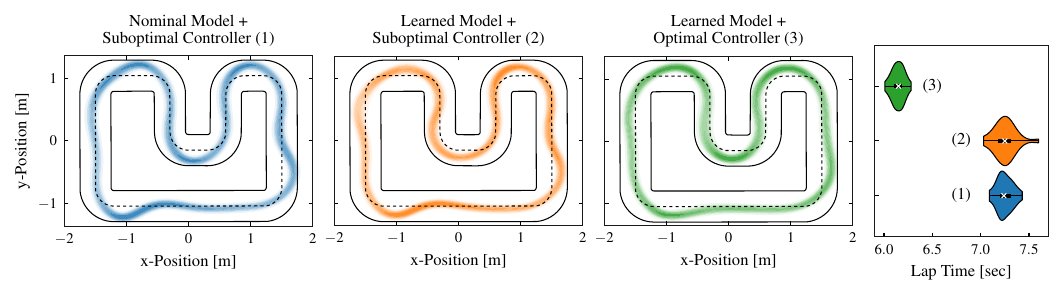}
    \vspace{-0.6cm}
    \caption{
        Three plots on the \revised{left-hand side}: density of the car's position across 20 laps for three different \revised{combinations of controller parameters and dynamic models}.
        Right-hand side: \revised{distributions of the} corresponding lap times for the three \revised{combinations}.
        \label{fig:position_density}
    }
\end{figure*}
\revised{The two questions that we aim to answer in this section are the following:
    How much does the residual model improve the predictive accuracy?
    And how much data is necessary to learn an accurate model?}

\cref{fig:learning_error_signal} shows the predicted error (solid line) and the corresponding uncertainty estimates (shaded region) for the lateral velocity and yaw rate, respectively.
\revised{Here,} the training data corresponds to two complete laps and the testing data to one lap.
\revised{The learned model is able to approximate the true error signal with high accuracy, although it only has two parameters per dimension. Further, the model does not overfit to the data but is able to generalize well.}
For a more quantitative analysis, \revised{we show a comparison of} our model with a \gls{gp} in terms of the RMSE in \cref{tab:prediction_error}.
\revised{The} uncertainty estimates for the prediction error correspond to the standard deviation from 50 models trained on independently sampled training (2 laps) and testing data (1 lap).
We note that learning with either model helps to drastically improve the predictive capabilities \revised{compared to} the nominal dynamics.
Notably, \gls{blr} with $\feature_{\text{Taylor}}$ even outperforms the \gls{gp} for the yaw rate, \revised{which indicates} that our model strikes a good balance between complexity and expressiveness.

\revised{To determine the right amount of training data for model learning, we have to consider the following trade-off:
    More data generally leads to a more reliable estimate of the regression coefficients.
    However, having a large data buffer prohibits the model from quickly adapting in the case of environmental changes.
    We therefore prefer to use as few data as possible without compromising on accuracy.
    \cref{fig:number_of_data} shows the regression coefficients' estimates for four contexts with an increasing amount of training data.
    With only few data ($<10$ seconds), the mean estimate of the coefficients fluctuates strongly.
    In contrast, the regression coefficients have mostly converged to stable values when using more than ten seconds of training data.
    For the remaining experiments, we therefore fix the size of the telemetry data buffer $\datatel$ to ten seconds, which corresponds to a little less than one and a half laps of the track.}
\begin{figure*}
    \centering
    \includegraphics[scale=0.95]{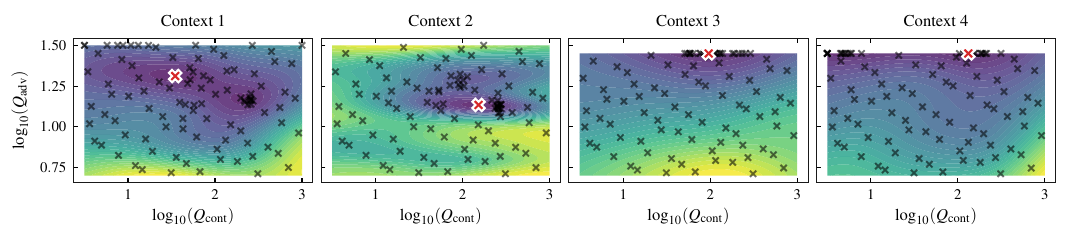}
    \vspace{-0.6cm}
    \caption{
        Response surfaces of the objective function \cref{eq:regularized_objective} for different contexts \revised{as a function of $Q_{\mathrm{cont}}$ and $Q_{\mathrm{adv}}$ (see \gls{mpcc} formulation \cref{eq:mpcc}).
        The evaluated controller configurations are denoted by the black crosses and the estimated optimum (minimum of the \gls{gp} posterior mean function) is shown as a red cross.}
    }
    \label{fig:laptime_response_surfaces}
\end{figure*}
\begin{figure*}
    \centering
    \includegraphics{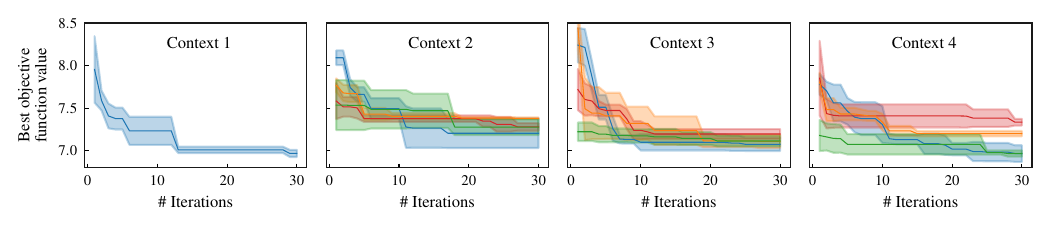}
    \vspace{-0.8cm}
    \caption{\revised{Learning progress of standard \gls{bo} (\protect\bostandardlegendentry), contextual \gls{bo} (\protect\bocontextuallegendentry), cumulative \gls{bo} (\protect\bocumulativelegendentry), and prior-mean \gls{bo} (\protect\bopriormeanlegendentry). The solid lines depict the mean and the shaded area corresponds to mean $\pm$ one standard deviation.}}
    \label{fig:contextual_bayesopt}
\end{figure*}

\revised{Next, we demonstrate that model learning alone is not sufficient for high-performance racing, but it allows for aggressively tuned controllers.
}
To this end, we drive 20 laps with three different settings and record the respective lap times:
\revised{The first setting} serves as the benchmark, for which the controller weights are not tuned and we only use the nominal model.
In \cref{fig:position_density}, we can see that the car reliably made it around the track albeit in a clearly suboptimal trajectory.
Especially after turns with high velocity (bottom left and right), the car repeatedly bumped into the track's boundaries (see also the accompanying video).
\revised{For the second setting,} we used the same controller weights, but account for errors in the dynamics by learning the residual model.
The car's trajectory is slightly improved and does not bounce against the boundaries anymore.
Interestingly, \revised{this} does not directly translate to a reduction \revised{in} the lap time.
\revised{In the third setting,} we changed the controller weights such that the driving behaviour is more aggressive\final{, which leads to a reduction in the lap time of more than a second. 
Moreover, the car's trajectory is greatly improved.
However, the trajectory does not fully converge to the optimal raceline, which we attribute to the \gls{mpcc} formulation that penalizes deviations from the centerline.}
Note that the combination of aggressive controller weights without model learning repeatedly lead to the car crashing into the track's boundary from which it could not recover.
We conclude that both accurate dynamics as well as a well-tuned controller are required for high-performance racing, justifying the joint optimization approach.

\vspace{-1mm}
\subsection{Contextual Controller Tuning}
\revised{In this section, we use the following objective function}
%
\begin{align}
    \label{eq:regularized_objective}
    J(\bovar) = T_{\mathrm{lap}} + \lambda \cdot \bar{\Delta}_{\mathrm{centerline}}, \quad \bovar = [Q_{\text{cont}}, Q_{\text{adv}}]^\top
\end{align}
with $T_{\mathrm{lap}}$ denoting the lap time, $\bar{\Delta}_{\mathrm{centerline}}$ is the average distance of the car to the centerline in centimeters, and \mbox{$\lambda$} is a parameter governing the trade-off between the two cost terms.
The regularizing term penalizes too aggressive behaviour \revised{by} the controller \revised{that results} in the car crashing due to cutting corners \revised{on the track}.
\final{As a consequence, no crash occured during the collection of the remaining presented results.}

\revised{With the following experiment, we confirm our hypothesis that the \gls{mpcc}'s parameters require adaptation under different environmental conditions}
For each context, \revised{we evaluate 100 different controllers}.
\revised{The first 70 controller parameters} are sampled from a low-discrepancy sequence to sufficiently explore the full parameter space.
\revised{The} remaining 30 \revised{parameter vectors} are chosen according to \revised{standard \gls{bo}}.
We show the respective \gls{gp} mean estimates of the objective function as well as its minimum (red cross) in \cref{fig:laptime_response_surfaces}.
The response surfaces vary considerable across the contexts and so do the optimal controller weights.
However, there are also some common features, e.g., a low advancing parameter $Q_{\text{adv}}$ in \cref{eq:mpcc} generally leads to cautious driving behaviour and therefore an increased lap time.

In the final experiment, we \revised{demonstrate} that \revised{our framework utilizes the information from previous runs more effectively compared to variants of the standard \gls{bo} algorithm.
    In particular, we consider the following three variants:
    \begin{itemize}
        \item Standard \gls{bo}: Serving as a naive benchmark, this variant does not include any information from previous runs, but starts from scratch for each context.
        \item Cumulative \gls{bo}: This variant accumulates all collected data without considering the contextual information.
        \item Prior-mean \gls{bo}: For this variant, we first learn a joint \gls{gp} from data of the previous contexts. The posterior mean then acts as the prior mean for the current \gls{gp}.
    \end{itemize}
    For all \gls{bo} methods, we employ the Mat\'ern kernel and infer the respective hyperparameters by evidence maximization.
    We evaluate contextual \gls{bo} and the three variants described above on the four different contexts and optimize} the weights $Q_{\text{adv}}$ and $Q_{\text{cont}}$ in the \gls{mpcc} formulation.
\revised{The results are depicted in \cref{fig:contextual_bayesopt}.}
For each context, we observe that standard \gls{bo} reliably finds the controller parameters that lead to the objective's minimum.
\revised{Recall} that each time the optimization is started from scratch such that the initial lap times are relatively high.
\revised{For context~2, all methods \revised{other} than standard \gls{bo} are able to exploit the additional information from the previous context.
    Here, contextual \gls{bo} does not exhibit a significant advantage over cumulative and prior-mean \gls{bo} because the optima for contexts~1~and~2 are similar.
    However, note that the response surface for context~3 is considerably different from the previous two contexts.
    Due to their heuristic nature, cumulative and prior-mean \gls{bo} converge even slower than standard \gls{bo}.
    In stark contrast, contextual \gls{bo} almost immediately obtains the optimal controller parameters.
    For the last context, we observe a similar picture: contextual \gls{bo} outperforms the non-contextual \gls{bo} variants.
    Cumulative and prior-mean \gls{bo} seem not to even find the optimal controller due to lack of exploration.
}


\addtolength{\textheight}{-0.5cm}   

\section{Conclusion}
\label{sec:conclusion}

We have presented a framework to combine model learning and controller tuning with application to high-performance autonomous racing.
In particular, we proposed \revised{encoding} the environmental condition via a residual dynamics model such that knowledge between different contexts can be shared to reduce the effort for controller tuning.
The benefits of the proposed approach have been demonstrated on a custom hardware platform with an extensive experimental evaluation.
Key for the method's success was the low-dimensional representation of the environment by means of the custom parametric model.
For more complex learning models such as neural networks, one could for example obtain a low-dimensional context via (variational) auto-encoders, which we leave for future research.



\footnotesize
\bibliographystyle{IEEEtran}
\bibliography{conference_names, bibliography}

\end{document}